# Agentic Temporal Graph of Reasoning with Multimodal Language Models: A Potential AI Aid to Healthcare

Susanta Mitra

AI Forum

susantamit@gmail.com

## Abstract

In the field of medical healthcare, LLMs have been used to improve the quality of medical tasks. Healthcare and medicine are multimodal disciplines that deals with multimodal data for reasoning and diagnosing multiple diseases. Integrating different modalities enables a more comprehensive understanding of the data and facilitates more sophisticated reasoning processes which is closer to the ways humans perceive the world. Although some multimodal reasoning models have emerged for reasoning complex tasks in scientific domains their applications in the healthcare domain remains limited and fall short in correct reasoning for diagnosis. The complex real-life medical reasonings demand dynamic pattern of thoughts and reasonings that are often non-linear or change in the flow of reasoning due to new or incorrect thoughts. To address the challenges of multimodal medical reasoning for correct diagnosis and assist the healthcare professionals a novel temporal graph-based reasoning process modelled through a directed graph has been proposed in the current work. It helps in accommodating dynamic changes in reasons through backtracking, refining the reasoning content and creating new or deleting existing reasons to reach the best recommendation or answer. The factual correctness of each reasoning step is additionally ensured by the online verification of extracted knowledge against extracted ground-truth sources. Again, consideration of the temporal aspects can add new dimensions to the reasoning and diagnosis process. Multimodal data at different time points can enable tracking and analysis of patient health and disease progression. Moreover, the proposed multi-agent temporal reasoning framework provides tasks distributions and cross-validation mechanism to further enhance the accuracy of reasoning outputs. Few basic experiments and analysis results justifies the novelty and practical utility of the proposed approach. More rigorous experiments with larger datasets and benchmarking to be done in future subject to the proper availability of real-life temporal medical reasoning data and comparative multimodal temporal reasoning models. An important point to note is that, whatever advancements are made with AI healthcare reasoning, the final verification presently remains with the human doctors, and the primary objective of AI agents should be to assist healthcare professionals with collaborative approaches to ease their tasks and not to replace them.

## 1 INTRODUCTION

Advancements in Artificial Intelligence (AI) and Large Language Models (LLM) have catalysed significant progress in healthcare research and clinical practice, especially in demonstrating significant progress and potential in medical data analysis (Sumon et al., 2025; Hernandez et al., 2022) and disease diagnosis (Nia et al., 2023). LLMs have played a crucial role in supporting specific applications, such as generating brief and accurate reports based on EHR, progress notes, doctor-patient conversations, and other medical text formats. In the field of medical healthcare, LLMs have been used to improve the quality of medical work in supporting specific applications, such as generating brief and accurate reports based on EHR, progress notes, doctor-patient conversations, and other medical text formats.

Recently, Multimodal Large Language Model (MLLM) (Tian et al., 2025; Li et al., 2023; Liu et al., 2023a; Zhang et al., 2023) has emerged as a promising research field that can perform multimodal tasks using LLMs as a brain. It has the ability to receive and reason multimodal information. The integration of different modalities enables a more comprehensive understanding of the data and facilitates more sophisticated reasoning processes. It is closer to the ways humans perceive the world and, therefore, can lead to the pursuit of developing Artificial General Intelligence (AGI). MLLMs that are capable of processing inputs from various modalities and delivering outputs across multiple modalities can be useful in healthcare applications. For example, NExT-GPT (Wu et al., 2024), an any-to-any MLLM, has been designed to seamlessly handle input and output in any combination of four modalities: text, image, video, and audio. Figure 1 presents the schematic overview of the NExT-GPT framework, consisting three main stages: encoding, LLM understanding and reasoning, and decoding. Although several pioneering studies have made preliminary attempts to apply MLLMs in the medical field, the works remain limited, still confined mainly to a physiological-level understanding rather than disease-level reasoning. Very few works, so far, have explored how to enhance the reasoning abilities of multimodal large language models in medical tasks, especially under complex healthcare scenarios. Moreover, none of the previous works have dealt with the temporal graph of reasoning on MLLM under a multi-agent framework. The current paper has focused on the preliminary work done on this to assist healthcare professionals in interpreting and analyzing medical data for effective decision-making and efficiently handling complex medical scenarios.

In summary, the key contributions of the current paper are as follows:

- *Agentic Temporal Graph of Reasons*, probably the first multi-agent temporal reasoning framework as proposed here, can be efficiently utilized for reasoning in the healthcare and medicine domain with the agent-based temporal reasoning process through multimodal language models. This framework can be significant for the development of AI-assisted healthcare system and provide good support to the real doctors and other healthcare professionals.
- The novel *temporal graph-based reasoning* modelled through a directed graph resembles complex human reasoning with the scope for accommodating dynamic changes in reasons through backtracking, refining the reasoning content and creating new or deleting existing reasons to reach the best recommendation or answer. The factual correctness of each reasoning step is additionally ensured by the online verification of extracted knowledge against extracted ground-truth sources. It enhances the reasoning capabilities and quality of answers to medical queries for better diagnosis and medical decision-making.
- Consideration of the temporal aspects can add new dimensions to the reasoning and diagnosis process. Multimodal temporal data can enable tracking and analysis of patient health and disease progression. The reasoning times created and stored through the temporal graph of the reasoning process during the inference /analysis period for a patient can be utilized to assess the patient's conditions, diagnosis, and treatment planning for that period. Moreover, reasoning graphs, final reasonings with initial and final times stored during different periods, can help to refer to the earlier archived reasonings for better understanding the current disease symptoms and diagnose accordingly.
- The multiple verifications in the explicit multi-agent temporal reasoning framework help to enhance the reasoning potential in complex healthcare scenarios and lead towards accurate answers. Final verification of the answers by the real doctors ensures the authenticity of the answers.

## 2  MLLM IN HEALTHCARE

Healthcare and medicine are multimodal disciplines. In health care, a health care professional makes a comprehensive approach for diagnosing and treating a patient. This often involves listening to the patient, reviewing their health records, helping develop treatment plans, medical text summarization, referral letter generation, completing insurance formalities, and analyzing laboratory test results. All of these activities are carried out over time. Apart from textual data, this multidimensional process also needs nontextual data types, such as *images, audio, video, sensory, omics, and temporal* data that play a crucial role in diagnosis, effective treatment planning, research, and patient care. Image data may include medical imaging (eg, X-rays, Magnetic Resonance Imaging or MRI, Computed Tomography or CT scans, positron emission tomography scans, and pathology slides) and electrophysiological data (eg, electrocardiography (ECG), electroencephalography (EEG), and electromyography), sensory data (eg, data from sensors of medical devices, such as pacemakers and continuous glucose monitors), videos (eg, recordings of surgeries, procedures, medical video question-answering and patient interactions), omics data (eg, genomics, proteomics, metabolomics, and transcriptomics), audio data (eg, recordings of patient interviews and heart and respiratory sounds) and temporal data (time-series data like real-time CGM, EEG and event data electronic health [EHR] records) (AlSaad et al., 2024).

The transition from MLLM to Medical Multimodal Large Language Models (Med-MLLMs) signifies a profound advancement in AI, particularly within the healthcare sector, though the development of LLMs and MLLMs in the medical field is still in its infancy, with many of their potential application scenarios remaining undefined. Moreover, they face a range of challenges, including hallucinations and a lack of up-to-date information, which significantly impede their practical use in healthcare settings (Xiao et al., 2024). However, the application of MLLMs in healthcare has gained significant attention due to their ability to process and integrate diverse data modalities. Current medical multimodal large language models are primarily based on general multimodal models and are further trained on specialized medical datasets (Mu et al., 2025). This has led to the emergence of Med-Flamingo (Moor et al., 2023), LLaVA-Med (Li et al., 2023a), Med-Palm M (Tu et al., 2024), BioMedGPT (Zhang et al., 2024), and MedTrinity-25M (Xie et al., 2024). Med-MLLM (Liu et al., 2023b), a medical multimodal large language model for future pandemics, has been designed for COVID-19 reporting, diagnosis, and prognosis. The primary applications of multimodal large language models in healthcare can be medical diagnosis, medical report generation, medical treatment, medical Visual Question Answering (VQA), clinical communication, clinical guidance, surgical assistance and medical education. A comprehensive survey and review of 330 recent papers in the areas of medical diagnosis, report generation and treatment carried out and specific examples provided to demonstrate the capabilities of MLLMs in these domains (Ye et al., 2025). The models (Wu et al., 2023; Hyland et al., 2024; Zhang et al., 2024a,b; Seyfioglu et al., 2025), demonstrate encouraging results in various tasks, including medical VQA, medical report generation, and diagnostic support, underscoring the considerable potential of MLLMs in healthcare scenarios. Subsequent research (Chen et al., 2024; Li et al., 2025b; Lin et al., 2025) developed refined training recipes as well as larger and higher-quality medical multimodal datasets. Med-Gemini (Saab et al., 2024), a family of models fine-tuned and specialized for medicine, has been built on the strengths of the Gemini models. As Gemini models are trained to accommodate textual input interleaved with a wide variety of other data modalities, they are known to excel in multimodal tasks.

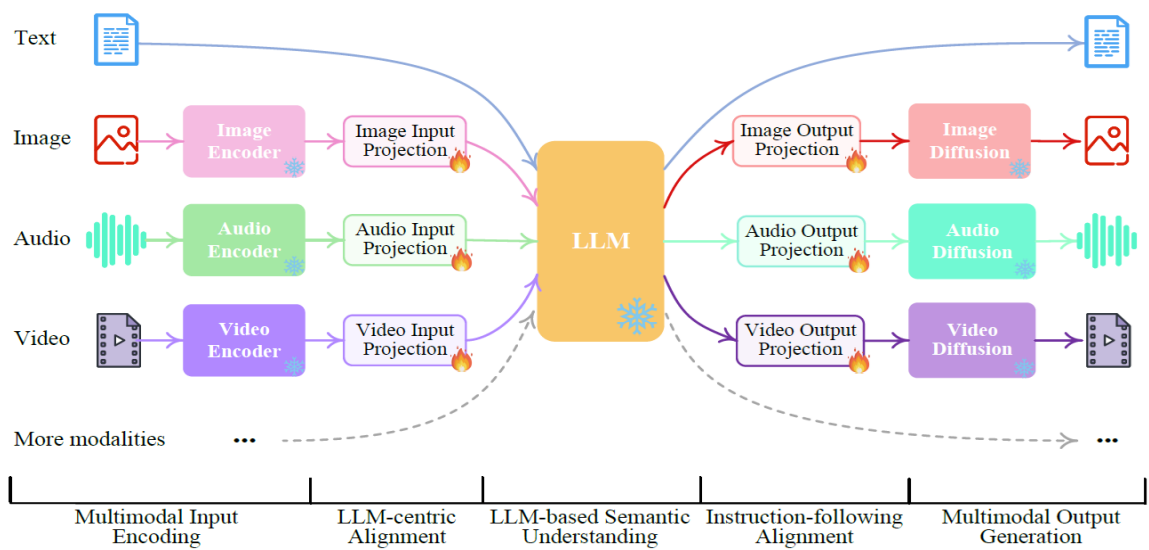

Figure 1. Schematic overview of the NExT-GPT framework

## 3 FOUNDATIONS OF REASONING

Reasoning is one of the fundamental intelligent behaviors of human beings, which requires understanding and analyzing given conditions and background knowledge to derive a new conclusion logically and rationally. Apart from logic rules, domain knowledge is also required to perform practical reasoning tasks. Hence, practical domains like healthcare or medical need substantial domain knowledge for reasoning. Dual-system theory suggests that human cognition operates through two modes: *System 1*, which is fast, automatic, and intuitive, enabling quick decisions with minimal effort, and *System 2*, which is slower, more analytical, and deliberate. Achieving human-level intelligence requires refining the transition from System 1 to System 2 reasoning (Li et al., 2025a). Foundational LLMs operate like System 1 reasoning, while Large Reasoning Model (LRM)s are designed to emulate the slower, more deliberate reasoning associated with System 2 thinking. Integrating dynamic reasoning strategies across different modalities (e.g., textual, image, and video data) allows for more efficient problem-solving by adjusting the cognitive complexity to match the nature of the inputs. (Qu et al., 2025). Extending slow-thinking capabilities to multimodal reasoning can be a promising direction, as real-world problems often involve multiple modalities. The applicability of multi-modal reasoning models will be significantly enhanced by developing such models that can integrate information from diverse sources while engaging in slow, deliberate reasoning (Pan et al., 2025a). The application of System 2 reasoning, like multimodal reasoning, can be practically useful in the domain of healthcare and medicine.

### 3.1 Reasoning Tasks

A concise overview of distinct categories of related reasoning approaches and tasks is provided here for background concepts. This comprehensive overview provides insights into the diverse landscape of reasoning tasks and approaches with references to the healthcare and medical field (Sun et al., 2023).

- Causal Reasoning
- Contextual Reasoning
- Visual Reasoning
- Multimodal Reasoning
- Retrieval Augmented Reasoning
- Agentic Reasoning
- Temporal Reasoning
- Medical Reasoning

**Causal Reasoning** - Causal Reasoning refers to the process of understanding and explaining cause-and-effect relationships between events, actions, or variables. Causal reasoning tasks can be categorized into causal discovery, effect inference, attribution, judgment, and other tasks. For example, in healthcare, causal reasoning helps determine the root causes of diseases and predict the effects of interventions, moving beyond mere correlations to understand true cause-and-effect relationships.

**Contextual Reasoning** - Contextual Reasoning refers to the ability to understand and interpret information by considering the surrounding context, enabling more nuanced and relevant responses, like how humans' reason. For example, in healthcare, contextual reasoning involves understanding and considering a patient's individual circumstances and factors beyond just their medical condition when making decisions about their care.  The professional might consider a patient's social support, financial situation, or cultural background when developing a treatment plan.

**Visual Reasoning** - Visual reasoning refers to the cognitive process of understanding, analyzing, and drawing conclusions from visual information. It involves the ability to perceive, interpret, and reason about visual stimuli such as images, scenes, or other visual representations. In the healthcare domain, reasoning plays a vital role in image analysis. Medical vision-language models (VLMs) combine computer vision (CV) and natural language processing (NLP) to analyze visual and textual medical data. VLMs exhibit promising reasoning capabilities on visual image analysis of various radiological tasks MRI, CT, and X-ray, through the unification of medical image analysis with explicit reasoning. (Pan et al., 2025b). Large-scale VLMs has spurred numerous domain-specific adaptations for healthcare, with systems such as LLaVA-Med (Wong et al., 2023) and HuatuoGPT-Vision (Chen et al., 2024) achieving impressive results in radiology VQA and related diagnostic tasks. MedCoT (Liu et al, 2024) improves medical visual question answering through a hierarchical expert system that culminates in a Mixture-of-Experts diagnosis.

**Multimodal Reasoning** - Multimodal Reasoning refers to the cognitive process of integrating and reasoning across multiple modalities of information, such as text, images, videos, and other sensory inputs, to enhance understanding and perform complex reasoning tasks. Multimodal reasoning aligns more closely with the way humans perceive the world. A systematic survey of Multiple Chain-of-Thought Reasoning (MCoT) reasoning is presented in (Wang et al, 2025a) that elucidates the relevant foundational concepts and definitions. Again, the application and evaluation of multimodal reasoning models in healthcare or medical domains, though limited, are increasing. A few recent notable works are (Mu et al., 2025; Zuo et al., 2025; Tang et al., 2025; LASA Team et al., 2025).

Multimodal reasoning tasks can be broadly categorized into image-text alignment, text-to-image generation, multimodal-to-text generation, and multimodal understanding. Multimodal foundation models mainly involve the following key techniques to approach reasoning tasks :

- *Multimodal In-Context Learning* - Conversation with an LLM /MLLM is carried out through inputs of various modalities and user text messages known as *Prompts* and LLM /MLLM responses with the final answer. In-Context Learning (ICL) is prompting LLMs with task-specific examples without additional explicit training to better interpret user intent and generate outputs aligned with expectations (Brown et al., 2020). ICL has been extended to more modalities, leading to Multimodal ICL. Multimodal ICL leverages contextual information to enhance multimodal reasoning.  Multimodal ICL can be used to solve various complex reasoning tasks. (Jin et al., 2024).

- *Multimodal Chain-of-Thought* – Chain-of-Thought (CoT) is prompting LLMs to reason step-by-step or break complex problems into logical steps (Wei et al., 2022). CoT reasoning emulates human problem-solving by decomposing complex tasks into a sequence of manageable sub-tasks, systematically constructing solutions. The intermediate reasoning steps or trajectories, termed the *rationale*, elucidate the logical progression underlying the model's conclusions. A single reasoning step in CoT is termed as *thought*. Extension of CoT to reason with multimodalities is termed as Multimodal Chain-of-Thought (MCoT). This augmentation broadens the scope of multi-step reasoning, enhancing its applicability to increasingly intricate scenarios. Reasoning with two or more multimodalities, (e.g., audio-visual) is termed as Cross-Modal CoT.

Various thought paradigms have emerged to enhance multimodal and multi-step reasoning. Based on the construction of thought generation during reasoning, the reasoning structures (Chu et al., 2023) or topologies (Besta et al., 2024a) have been categorized into chain, tree, and graph types. In these topologies, thoughts are represented as nodes, with edges indicating dependencies between them. Chain topologies facilitate linear and sequential thought generation, progressively converging toward the final answer, while tree or graph structures can generate multiple child nodes from a single parent and enable exploration and backtracking within the reasoning process. Graph structures can further introduce cycles and a single node can have multiple parent nodes, which facilitates aggregation among multiple nodes (Wang et al., 2025a). Hence, based on these topologies, thoughts can be categorized as follows.

- *Self-Consistency with CoT (CoT-SC)* -Self-Consistency with CoT can generalize CoT into multiple CoTs by generating multiple independent reasoning chains or paths instead of a greedy one and selecting the most consistent one to provide the best output (Wang et al., 2023).
- *Tree of Thoughts* - Tree-of-Thoughts (ToT) (Yao et al., 2024) extends CoT-SC reasoning by exploring multiple reasoning paths in a tree-like structure and provides the scope of "backtracking" or local exploration of a path.
- *Graph of Thoughts* - Graph-of-Thoughts (GoT) generalizes CoT and ToT to handle a complex network of thoughts like *recurrence* that brings LLM reasoning closer to human thoughts. It enables the arbitrary connection of thoughts and their refinements. Like Multimodal-CoT (Zhang et al., 2023), (Yao et al., 2023) modelled GoT reasoning as a two-stage framework, generating rationales first and then producing the final answer. To bring LLM reasoning closer to human thinking or brain mechanisms, such as recurrence, a GoT model (Besta et al., 2024b) was proposed to represent complex reasoning networks through a Directed Acyclic Graph. Again, Adaptive Graph of Thoughts (AGoT) (Pandey et al., 2025) have attempted to unify the strengths of chain, tree and graph into a cohesive network that reduces computational overhead by dynamically decomposing the complex queries into structured subproblems.

**Retrieval-Augmented Reasoning** - Retrieval-Augmented Reasoning (RAR) combines Retrieval-Augmented Generation (RAG) techniques with reasoning to enhance the capabilities of language models, particularly for complex, knowledge-intensive tasks. RAG enhances LLM reasoning by incorporating external knowledge. RAR incorporates reasoning abilities, such as logical deduction, abductive reasoning, and problem-solving, to process the retrieved information and generate answers or insights. Retrieval-Augmented Reasoning Modeling (RARE), a novel paradigm proposed in (Wang et al., 2025c) decouples knowledge storage from reasoning optimization. RARE externalizes domain knowledge to retrievable sources and internalizes domain-specific reasoning patterns during training and inference. It combines domain knowledge and reasoning.

**Agentic Reasoning** - Agentic Reasoning is a crucial capability for AI Agents, referring to a cognitive process that integrates perception, action, and interaction with the physical or simulated environment to support reasoning and problem-solving. This enables AI agents to think, learn, and make decisions in real-time, adjusting to changing environments and transforming knowledge into action. The key components of Agentic Reasoning systems are Introspective Reasoning, Extrospective Reasoning, Embodied Reasoning, and Multiagent Reasoning. These capabilities are essential for creating intelligent systems that can successfully operate in complex and dynamic environments, seamlessly interact with humans, and engage in cooperative or competitive scenarios with other agents (Sun et al., 2023).

*Multi-agent* reasoning refers to the cognitive process by which multiple autonomous agents or entities engage in reasoning, decision-making, and communication within a shared environment or context. Compared with reasoning with a single agent, it involves the ability of individual agents to perceive, interpret, and reason about the actions, goals, beliefs, and intentions of other agents, and to adjust their own behaviors accordingly. Recent studies have introduced the concept of multi-agent debate as a promising method to elevate reasoning abilities and ensure factual accuracy across diverse scenarios.

**Temporal Reasoning** - Temporal Reasoning in Language Models (LM) refers to the ability of LMs to understand, represent, and reason about time and its relationship to events and other entities. It enables AI to model how data evolves, which is critical for predicting future states, analyzing historical trends, or responding to real-time inputs.

Temporal reasoning is involved in several major medical reasoning tasks. It can be applied to support time-oriented decision making for each one of these fundamental stages of patient care : (1) *Prevention*, which predicts risk factors using methods such as time series analysis; (2) *Diagnosis*, which discovers patterns of temporal evolution of diagnostic evidence ; (3) *Treatment*, which administers a therapeutic protocol (e.g., chemotherapy) over several predetermined periods; and (4) *Prognosis*, which forecasts the effects of health care on the problems of the patient over a period of time (Zhou and Hripcsak, 2007). Multimodal temporal medical reasoning involves using AI to analyze various multimodal data types like text, images, audio, video and time-series data to understand and reason about the temporal evolution of patient conditions and medical events. It is crucial for understanding patient histories, tracking disease progression, and making informed treatment decisions. This can help doctors make accurate diagnoses by understanding the sequence of symptoms, their timings, and tracking patient progress over time by analyzing multimodal data to assess treatment effectiveness.

**Medical Reasoning** – Healthcare or medical tasks often need reasoning, especially for complex cases. For example, clinicians reason the potential causes of a patient's symptoms and then advise which examinations to take and what treatment is best following the diagnosis. Foundation models can conduct expert-level reasoning in the context of medicine Med PaLM 2 (Singhal et al., 2025). Here, human expert annotation is used, where clinicians provide consensus and rationale for reasoning paths. (Chen et al., 2024) have introduced two-stage sophisticated verifiable approaches for medical reasoning and Huatuo GPT-o1, a medical LLM capable of complex reasoning. However, the model may sometimes generate factually incorrect explanations that can lead to plausible hallucinations. The problem due to this type of surface-level explanations can be addressed through impose external structure, like a medical knowledge graph (Wu et al., 2025a). For reasoning quality assessments, MedR-Bench (Qiu et al., 2025) introduces a "Reasoning Evaluator", an automated tool that scores free-text clinical reasoning responses along multiple dimensions: efficiency, actuality, and completeness. A recent systematic review of enhancement techniques and applications of medical reasoning is provided in (Wang et al., 2025b).

## 3.2 Reasoning Techniques

To explore the reasoning capabilities of multimodal language models, the research community has investigated the reasoning capabilities of language models in multimodal contexts and developed several methods and strategies. The reasoning processes can adopt either text-only or multimodal rationales. Moreover, to equip MLLMs with robust medical reasoning, researchers have developed a suite of techniques. The main approaches considered in this work can be broadly categorized into two main stages: training-time techniques and test-time techniques.

### 3.2.1 Training-time Techniques

Training-time techniques imbue medical logic directly into the model's parameters to develop the reasoning capabilities. These techniques fundamentally alter a model's internal weights to build foundational reasoning power. The two main methods of training-time techniques are Supervised Fine-Tuning (SFT) and Reinforcement Learning (RL).

**Supervised Fine-Tuning** - Supervised Fine-Tuning (SFT) provides training on data containing explicit reasoning chains that force the model to learn the 'why' and 'how' of diagnosis. It includes few strategies which are crucial for multimodal reasoning to be useful for diagnosis.

**Reinforcement Learning** - Reinforcement Learning (RL) is the alignment capability to fit the nuanced goals of medical practice: safety, accuracy, and efficiency. RL can act as a powerful engine for complex medical reasoning. A spectrum of feedback strategies, from holistic human judgment to granular, and awarding rewards have been developed for defining clinical reasoning. This includes RL with Human Feedback (**RLHF**) (Singhal et al., 2025), RL with AI Feedback (**RLAIF**) (Chen, et al., 2024) and optimization against objective, quantitative metrics using **Structured Rewards** like Reinforcement Learning with Verifiable Rewards (RLVR). RLVR is an approach gaining importance in medical reasoning. It leverages automated verifiers to provide objective feedback to language models, guiding them towards more accurate and logically sound medical reasoning, especially in complex scenarios where human feedback might be subjective or inconsistent. A few recent works are (Niu et al., 2025; Xu et al., 2025; Lai et al., 2025; Su et al., 2025; Zhang et al., 2025a; Zhang et al., 2025b).

### 3.2.2 Test-time Techniques

Test-time techniques are on-the-fly mechanisms for dynamic reasoning of the pre-trained language models. Unlike training-time techniques, test-time techniques offer a flexible and low-cost way to steer the reasoning of pre-trained models. The three broad methods of training-time techniques are: Rational Generation, Multi-agent Reasoning, and Test-time Scaling.

**Rationale Generation** - The rationales need to be constructed first with one of the three distinct categories of methodologies for MCoT reasoning: Prompt-based, Plan-based and Knowledge Enhanced (Wang et al., 2025a).

- *Prompt-based* - Prompt-based reasoning method employs carefully designed prompts, including instructions or in-context demonstrations, to guide models in generating rationales during test-time or inference, typically in zero-shot or few-shot settings.
- *Plan-based* - Plan-based reasoning method enables models to dynamically explore and refine thoughts during the reasoning process. Unlike prompt-based methods with their linear, example-driven inference, plan-based variants enable models to traverse multiple reasoning pathways, enhancing adaptability and problem-solving depth. Plan-based MCoT Reasoning Process is illustrated through Figure 2 as borrowed from (Wang et al., 2025a).

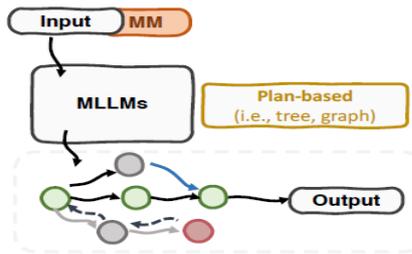

Figure 2. Plan-based MCoT Reasoning Process

- Knowledge-Enhanced Reasoning (Wang et al., 2025b) method utilizes Retrieval-Augmented Reasoning (RAR) to address the critical issues of hallucination and outdated knowledge by grounding the model's parametric memory in verifiable, external facts. Here, the model first queries a medical database or text corpus for relevant information, then integrates this retrieved text into its context to generate a factually grounded answer for the medical question or a query (Jeong et al., 2024; Zhan et al., 2024).

**Multi-agent Reasoning -** Multi-agent Reasoning is based on a distributed, specialized cognitive architecture where the LLM acts as an "orchestrator". This approach decomposes complex problems into tasks solved by multiple collaborating agents with their individual reasoning. (Kim et al., 2024) attempted a novel multi-agent framework, named Medical Decision-making Agents (MDAgents), that helps to address complex medical tasks by automatically assigning a collaboration structure to a team of LLMs. The assigned solo or group collaboration structure is tailored to the medical task at hand; a simple emulation inspired by the way real-world medical decision-making processes are adapted to tasks of different complexities. The strategy is applied to Medical Decision-Making (MDM), a task that requires teamwork and should benefit from multi-agent collaboration. MDM involves interpreting complex and multi-modal data, such as imaging, electronic health records (EHR), physiological signals, and genetic information, while rapidly integrating new medical research into clinical practice. MEDAGENTS (Tang et al. 2024), a multi-disciplinary collaboration framework for the medical domain, leverages LLM-based agents in a role-playing setting that participate in a collaborative multi-round discussion, thereby enhancing LLM proficiency and reasoning capabilities. Again, (Peng et al., 2025) have proposed a tree reasoning structure to store the reasoning paths and the corresponding clinical evidence. Moreover, a multi-agent framework was designed to handle complex medical scenarios and make decisions through multiple agents.

**Test-time Scaling -** Test-time or Inference-time scaling has emerged as a powerful technique for enhancing the reasoning capabilities of large language models (LLMs). Test-time scaling is the practice of using extra computational resources during inference to improve model performance. Instead of solely relying on pre-training and post-training adjustments, test-time scaling allows the model to dynamically allocate resources and refine its reasoning process during actual usage. "Slow Thinking" in LLMs, based on System 2, is an optimal scaling test-time computation during inference that may outperform scaling model parameters in efficiency. Test-time scaling can significantly improve medical reasoning in large language models without requiring extensive fine-tuning. A comprehensive investigation of test-time scaling for medical reasoning is provided in (Huang et al., 2025a, Huang et al., 2025b), which is a simple yet effective approaches that increase a model's medical reasoning capability during inference.

## 4   PROPOSED FRAMEWORK

The capabilities of Language Models have been enhanced through the domain-specialized open-source models that were pre-trained or fine-tuned on medical datasets or utilizing general-purpose MLLMs like GPT-4o with their intrinsic or latent medical knowledge. However, these are not sufficient for medical reasoning in real-life complicated medical cases that are collaborative and multidisciplinary in nature, with the usual involvement of multimodal data. So, a multi-agent system with different expertise is needed to collaborate for better reasoning, cross-verifications, and producing more accurate results.

Inspired primarily by the multi-agent frameworks of (Tang et al., 2024; Kim et al., 2024; Zuo et al., 2025; Yan et al., 2024; Wang et al., 2025d; Peng et al., 2025), a novel *multi-agent temporal graph-based reasoning* is proposed here for effective diagnosis and medical decision making. Depending upon the complexities of the input queries and the severity of cases, scope is there for collaborative multi-round discussion among a multidisciplinary or multidomain team of experts, thereby enhancing multimodal language model proficiency and multimodal reasoning capabilities. Role-playing in a multi-agent framework allows the model to explicitly reason with more individual professional and academic domain knowledge in addition to the knowledge base created from external knowledge via cloud (Zhao et al. 2024). Again, a novel concept of multimodal temporal graph of reasons is proposed in Section 5 to reason out the temporal evolution of patient conditions and carried out medical events for each agent. Each agent performs the fundamental tasks like patient observation and diagnosis in temporal order and can generate a directed graph of reasons through temporal medical reasoning during inference. The relevant answer to the input query and the final reasoning path with initial and terminal node creation times are stored by the agent for future comparison and analysis. The graphs with the reasonings created at different time points during an investigation /analysis period can be utilized to assess the patients' conditions, diagnosis and treatment planning for that period. Next, the expert agents share their individual information through multiple rounds of discussion and analyze different diagnostic conclusions from their respective professional perspectives. This multi-agent verification mechanism corrects and updates the diagnostic paths of different agents. The diagnostic paths of all agents are then summarized to provide the final answers to the medical problems of the patients. Finally, the final answers are manually verified by the Primary Doctor to ensure that reasoning content is supported by authoritative medical evidence or reference ground-truth reasoning. The final answer with the diagnosis is provided by the Primary Doctor to the patient if the answer is alright. Multiple levels of verifications of the final answer can ensure the accuracy of reasoning and minimizes the chances of hallucinations.

For a given medical query as an input or a prompt, the framework iteratively performs reasoning through the following six stages:

**Query Assessment** - Assessment of the medical query from a Patient is usually done by a Primary Doctor (PD) Agent or General Medical Practitioner (GMP) Agent to understand the severity of cases and corresponding types of medical expertise, if needed for diagnosis and treatment.

**Domain Experts Activation** – Relevant Expert(s) Agent(s) from various disciplines are activated based on the query evaluation and the severity. A GMP may handle the cases with mild or low severity without any consultations or support from other experts. For moderate cases, GMP recommends for relevant expert consultations to the patient to facilitate further medical investigations and analysis before the final diagnosis. For severe cases like multiple chronic conditions, complicated surgical or trauma cases and conditions that demand complex decision making involving different expert teams and departments, GMP usually recommends hospitalization for thorough investigation and treatments accordingly.

**Individual Expert Analysis** – Expert(s) carry out their own analyses through relevant medical investigations like pathological, imagological examinations, if necessary, followed by temporal reasoning and generating a final answer to the query. Temporal Reasonings can help experts to study and analyse patients' symptoms over a period of time and compare with reasonings and diagnoses during past periods for existing patients. The process involves generating individual Temporal Graph of Reasons to represent the reasoning rationales as detailed in Section 5.

**Analysis Synthesis** – For severe cases, multiple teams of experts from relevant disciplines carry out separate investigations and prepare the team report based on the analysis. A summarized or synthesized report is generated by extracting key knowledge and total analysis based on the individual team reports.

**Collaborative Consultation** – Multiple rounds of discussions or consultations on the summarized report can be carried out among lead team experts and ultimately render a summary report that is recognized by all experts and approved by the Primary Doctor agent.

**Decision Making** – A final decision is derived from the unanimous summary report by the Primary Doctor to provide a well-informed answer to the medical query. This final decision can be considered as the final diagnosis based on which the treatment plan is prepared and communicated to the patient agent.

An overview of agentic temporal reasoning framework is provided in Figure 3. The image icons for multimodal inputs, knowledge and cloud as shown at the top of the diagram are equally applicable for 'Mild' and 'Moderate' cases but not repeated for those cases to avoid clumsiness of the diagram.

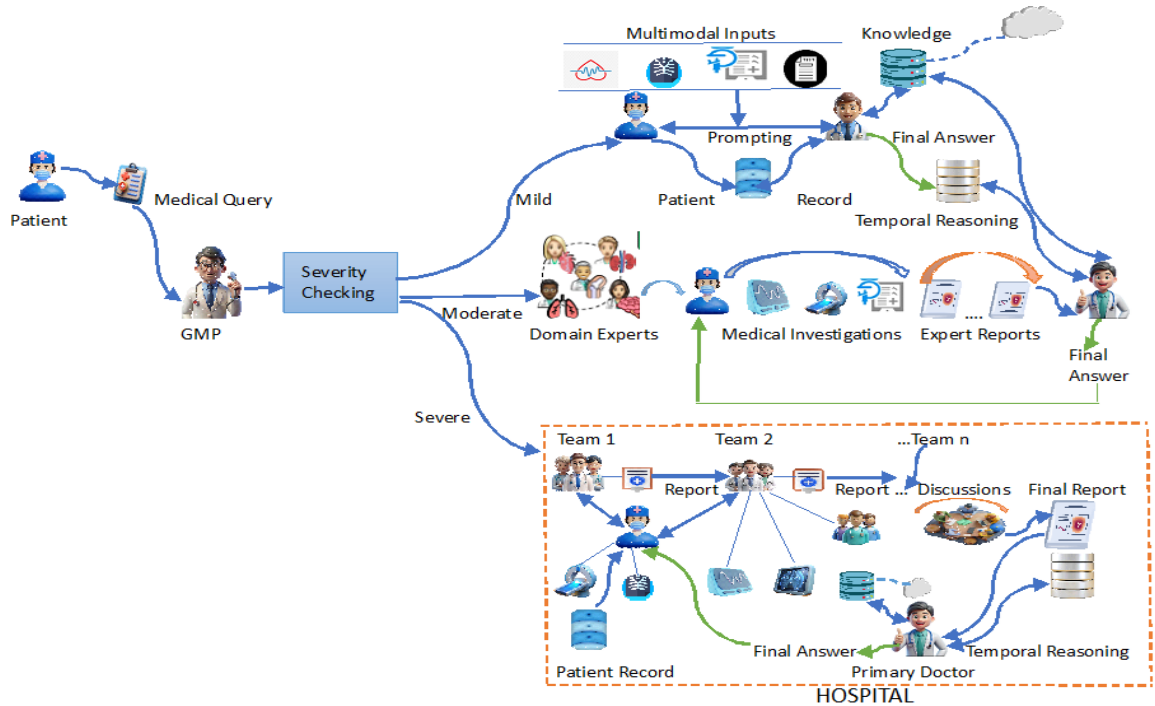

Figure 3. Agentic Temporal Reasoning Framework

## 5 METHODOLOGY

The methodologies followed in the current approach mainly consist of :

- Reasoning Data Curation
- Model Training
- Reasoning Generation for Multi-Agent Reasoning

### 5.1 Reasoning Data Curation

Medical reasoning or cognition refers to studies of cognitive processes, such as perception, comprehension, decision making, and problem solving in medical practice itself or tasks representative of medical practice. Medical reasoning examines the cognitive (thought) processes involved in making medical decisions. It involves an inferential process for making diagnostic or therapeutic decisions or understanding the pathology of a disease process (Patel et al., 2004). A key challenge in medical reasoning is verifying the thought process, which often lacks clear steps. The data curation pipeline includes the following steps for multimodal medical data:

**Data Collection**: In this phase, data are carefully selected and collected with more difficult and challenging problems that require expert-level multimodal understanding capability of foundation models, nuanced perception, and domain-specific knowledge to perform step-by-step reasoning to derive the solution for complex medical reasoning tasks. Inspired primarily by MedXpertQA (Zuo et al., 2025), MedAgentsBench (Tang et al., 2025), ReasonMed (Sun et al.,2025), MultiCaRe (MultiCaRe, 2025), MedTrinity-25M (Xie et al., 2025), MMIST-ccRCC (Mota et al., 2024), PubMedVision (Chen et al., 2024a), QuiltLlaVA Instruct (Seyfioglu et al., 2025), a synthetic multimodal medical reasoning dataset was carefully constructed from the mentioned publicly available datasets.

**Data Filtering**: To further control the quality of data, the steps followed are:

i. Identification and elimination of potential duplicate problems.
ii. Categorization of problems into Mild, Moderate and Severe and selecting the challenging questions.
iii. Transforming to verifiable medical problems by reformatting the closed set question to open-ended problems (Chen et al., 2024b)

iv.   Typographic, factual inaccuracies, missing information and format checking of the problems.

v.   Final review of the problems by a few medical experts.

## 5.2   Model Training

A two-phase training framework for multimodal temporal reasoning is proposed here. First, in Phase 1, Supervised Fine-Tuning (SFT) activates the model's reasoning potential using separate high-quality domain-specific multimodal (text, visual and audio) reasoning datasets. Subsequently, in Phase 2, Reinforcement Learning with Verifiable Reward (RLVR) and Group Relative Policy Optimization (GRPO) enhances reasoning capabilities further as it has been increasingly adopted for training MLLMs for reasoning enhancements. Applications of RLVR with GRPO in medical domains for reasoning (LASA Team et al., 2025) and especially for Visual Language Model (VLM) have recently been done in (Lai et al., 2025; Pan et al., 2025; Tan et al., 2025) for improving generalizability, interpretability, and reliability, and overcoming a few limitations of Supervised Fine-Tuning (SFT). An attempt has been made here to preliminarily explore the potential of applying reinforcement learning with verifiable rewards paradigm to enhance medical reasoning ability.

**Phase 1: SFT-based Training** - In this phase, Supervised Fine-Tuning (SFT) is employed on the curated multimodal dataset to fine-tune MLLM. Here, the model deeply explores and refines its reasoning before answering. The training objective maximizes the likelihood of generating both rf and af given (i,q) is given by

$$\mathcal{L}_{\text{SFT}}(\theta) = - \mathbf{E}_{(i,q,r,a) \sim D} \sum_{t=1}^{T} \quad \log \pi_\theta \left(s \mid i, q, s_{<t}\right),$$

where $\theta$ is the model parameter, i represents the set of one or more multimodal inputs, q is the query, r is the reasoning steps, a is the final answer, $D$ is the curated multimodal dataset, $s$ is the concatenated sequence of r, a and $\pi_\theta$ is the model's token distribution. The output model $\pi_{\text{SFT}}$ is used for initialization for Phase 2 and serves as a solid foundation for reinforcement learning.

**Phase 2: RL-based Reasoning Enhancement** - The verifiers in RLVR provide clear, objective rewards (or penalties) based on whether the model's reasoning steps and final conclusions are logically sound and factually correct. The verifiable rewards help the model learn to prioritize accurate and reliable reasoning paths and generalize better to new and unseen medical scenarios. RL-based Group Relative Policy Optimization (GRPO) calculates relative advantages by comparing rewards within a group of sampled actions rather than a value function, reducing computational overhead and simplifying optimization. Moreover, it employs a set of fixed rules as the reward signal instead of a learned reward model. These optimizations make GRPO 50% more resource- and computation-efficient than Proximity Policy Optimization (PPO) (Shao et al., 2024; Schulman et al., 2017). Reason-RFT (Tan et al., 2025) introduces a two-phase training framework for visual reasoning: (1) Supervised Fine-Tuning (SFT) with curated Chain-of-Thought (CoT) and (2) Group Relative Policy Optimization (GRPO)-based reinforcement learning that generates multiple reasoning response pairs, significantly enhancing generalization in visual reasoning tasks. The application of GRPO using a reward function, the Jaccard Reward, to enhance the accuracy of multi-disease prediction through multimodal tasks has been proposed in (Zhang et al., 2025b).

Let $\pi_{\theta(\text{init})}$ be the initial policy model which is usually the initial SFT model, $\pi_{\theta(\text{old})}$ and $\pi_\theta$ be the old and current(new) policy models respectively, $\pi_{\text{ref}}$ is the reference policy which in practice the frozen base MLLM, such that $\pi_\theta \leftarrow \pi_{\theta(\text{init})}$ for policy initialization and $\pi_{\text{ref}} \leftarrow \pi_\theta$ before each iteration of GRPO sampling process. Let MRL be the dataset identified earlier for RL and P(Q) denote the question set from MRL for training such that any question q ε P(Q). The steps followed for RLVR supervision with GRPO optimization are formally presented below.

i.   For each input mode (e, q), where e is the encoding of the input mode and q is the textual encoding of the question and during each iteration, old policy model $\pi_{\theta(\text{old})}$ is updated with the current policy model $\pi_\theta$ followed by sampling a group G of outputs, $\{a_1, a_2, \ldots, a_G\}$ from the old policy model $\pi_{\theta(\text{old})}$. The sampling process can be expressed as

$$\{a_i\} \sim \pi_{\theta(\text{old})} \left(a_i \mid e, q\right), \quad \text{for } i = 1, 2, \ldots G$$

ii.   Each sampled output ai is assigned a reward based on the verification criteria, resulting in rewards or a reward set R{$r_1$, $r_2$, ...$r_i$, ...,$r_G$}. A reward $\mathcal{R}$ can have two components for an input mode apart from text mode: Format Reward ($\mathcal{R}_{\text{Format}}$) and Accuracy Reward ($\mathcal{R}_{\text{Accuracy}}$), such that $\mathcal{R} = \mathcal{R}_{\text{Format}} + \mathcal{R}_{\text{Accuracy}}$

*Format Reward* need to adhere to a strict format by following a pre-defined template with <think> and <answer> tags for a structured and interpretable responses during reasoning process. A reward score of '1' is given for correct adherence to format and '0' otherwise.

Format :     \<think\> {reasoning…} \</think\>
             \<answer\> {final answer} \</answer\>

$$\mathcal{R}_{\text{Format}} = \begin{cases} 1, & \textit{if correct adherence to the format} \\ 0, & \textit{otherwise} \end{cases}$$

*Accuracy Reward* is a rule-based reward that checks if the actual answer matches with the ground truth. A reward score of '1' is given when the results match and '0', otherwise.

$$\mathcal{R}_{\text{Accuracy}} = \begin{cases} 1, & \textit{if answer matches with the ground truth} \\ 0, & \textit{otherwise} \end{cases}$$

iii.     Rewards are normalized at the end of each output $a_i$ to estimate relative advantages {$A_1$, $A_2$,……., $A_G$), where

$$A_i = \frac{r_i - \text{mean}(R)}{\text{standard deviation } (R)}$$

The policy is updated based on the positive advantages, and the policy model is optimized by minimizing the following GRPO objective functions, along with the minimization of KL divergence between the trained policy and the reference policy. The KL divergence is estimated with the unbiased estimator, $D_{\text{KL}}$ [$\pi_\theta \parallel \pi_{\text{ref}}$] as provided in (Schulman, 2020). The GRPO objective function can be expressed by an equation similar to Equation (1) in (Lai et al., 2025).

iv.     The final optimized policy model is returned and saved.

### 5.3    Reasoning Generation for Multi-Agent Reasoning

The healthcare or medical domain often involves complex reasoning. In the real world, healthcare professionals need to be very careful during diagnosis. Such a life-critical field demands meticulous reasoning to ensure the most reliable answers. Moreover, reasoning depends heavily on domain-specific knowledge that is evolving also and not always available within pre-trained models, necessitating fact-based retrieval from external sources. Lack of knowledge or incorrect knowledge can lead to incorrect medical reasoning and inaccurate answers during inference time. The factual correctness of each reasoning step is additionally verified by verifying extracted knowledge against extracted ground-truth sources. The medical reasoning process should be verifiable and checked against the ground-truth answer for correctness. The verification is done on the accuracy of the reasoning data before generating the final answer to ensure quality. For reasoning, a plan-based rationale generation process is adopted in the current work with a dynamic, graph-based inference framework. Here, multiple reasoning pathways help in Exploring New Paths, Verification, Correction, and Backtracking until a final correct answer is derived. A directed temporal graph is created during the reasoning process as discussed in detail in the next section. This graph is generated during inference time, ensuring that it predominantly consists of correct reasoning paths, while invalid paths are rejected.

### 5.3.1 Graph of Reasons

It has long been recognized that the human thought process is far more complex and non-linear than could be captured by a simple, sequential chain of thoughts. This non-linear, jumping thought process is a hallmark of human creativity, reasoning, and problem-solving abilities that need to be modeled properly. A graph is a powerful non-linear structure that is able to capture the rich, non-sequential nature of human thoughts and allows for more realistic and logical modeling of reasoning processes. To bring LLM reasoning closer to human thinking or brain mechanisms, such as, a GoT model was proposed to represent complex reasoning networks through a Directed Acyclic Graph recurrence (Besta et al., 2024). Again, Adaptive Graph of Thoughts (AGoT) (Pandey et al. 2025) have attempted to unify the strengths of chain, tree and graph into a cohesive network that reduces computational overhead by dynamically decomposing the complex queries into structured subproblems. However, the reasoning processes in these models are mainly confined to text space reasoning without any multimodal or temporal considerations, and the model decisions cannot take place in a feature space with multimodal input parameters in the medical domain. The current paper has attempted a novel approach by incorporating *temporal graph reasoning* through a temporal graph-based multi-agent reasoning framework to enhance the reasoning capabilities of MLLMs in the healthcare and medical domain. The reasoning graphs created during different time intervals can be help to understand the dynamics of medical phenomena for better treatment planning and patient care. Moreover, the reasoning times required by different types of medical queries posed to the agents can be assessed for analysis. In the present paper, all the reasoning over time, along with its relationships for each context, is represented through a directed temporal graph, where each reason is represented as a node or vertex that covers the reason for the node, the answer or response to the reason, and the corresponding node creation time.

Formally, it can be said that a *Temporal Graph of Reasons, G* is a directed graph to model the reasoning process of a MLLM over a time $T$, where, $G = (V, E, T)$; $V = \{v_1, v_2, \ldots, v_i, \ldots, v_n\}$ is a set of vertices, $|V| = n$ is the number of vertices, $T = \{t_1, t_2, \ldots, t_i, \ldots, t_n\}$ is the set of corresponding discrete time instances at which the vertices are created and $E \subseteq V \times V$ is a set of temporal edges, $|E|$ is the number of temporal edges. Any two arbitrary nodes in $G$ may either be connected by one or more distinct edges and the collection of such edges is termed a *reasoning path*.

Let $Q$ be the set of questions or input queries, $I$ be the set of different types of multimodal inputs (text, image, audio, video, temporal etc), $A$ be the set of ground-true answers, $R$ be the set of all possible reasons, $S$ be the set of all reasoning strategies and $M$ be the set of verified medical problems covering tuples with medical problems /questions and the corresponding ground-true answers such that $M = \{(q, a^*)\}$, where $q \in Q$ and $a^* \in A$. For a given input text query or a question and a multimodal input, there can be different reasonings, where each reasoning set $R_i$ consists of a collection of reasons $r_j$ such that each $r_j \in R_i$ and $R = \cup R_i$. We can consider a path of reasons $p$ to be a collection of such reasons where $p = \{r_0, r_1, r_2, \ldots, r_{j-1}, r_j, \ldots r_n\}$. Again, each reason can be considered as a collection of tokens, and each $R_i$ leads to an output /answer $a_k$ such that $a_k \in A$. So, any arbitrary node, $v_j$, corresponding to a reason can be represented as a tuple $v_j = (r_j, a_j, t_j)$ where $r_j \in R$, $a_j \in A$ and $t_j \in T$.

### 5.3.1.1 Graph Reasoning Process

Temporal Graph of Reasons, $G$ is generated layer by layer through applications of reason transformations based on the chosen reasoning strategies by the trained MLLM. Each such transformation is a function that maps input to a new reason and answer at a time instant $t'$. If $\Gamma$ be a transformation function that transforms G to G' at $t'$, then $G' = \Gamma (G)$, where $G' = (V', E', T')$.

The transformation is initially applied on the input query or the question to generate a reason with the answer. The answer is verified with the ground-true answer. If the answer is incorrect then MLLM iteratively refines the model by selecting suitable reasoning strategies as defined here till the answer matches the ground truth answer.

**New Reason Exploration** – MLLM explores new reason $r_n$ distinct from the prior reasons and may subsequently explore more new reasons. New reasons as nodes are inserted into G to generate G'.

**Reason Content Refinement** – A current reason is refined through the modification of the content of the reason. This is represented through a loop in the graph that indicates an iterated reason with the same node connection as the original one.

**Backtracking** – MLLM recognizes reasoning path as suboptimal and revisits a previous reasoning node and continue reasoning from there to explore alternative possible reasoning paths. Self-Backtracking techniques equip the language models to backtrack during both training and inference (Yang et al., 2025).

**Generation of Reasons** – MLLM generates one or more new reasons based on the existing single reason. This can give rise to a *cap-like* structure where, say, three vertices $v_2$, $v_3$ and $v_4$ are linked to a root vertex $v_1$. The outdegree of the root vertex should be more than 1 and here outdegree($v_1$) = 3.

**Merging of Reasons** – MLLM merges or aggregates reasons into a new reason. This can give rise to a *cone-like* structure where, say, two vertices $v_1$ and $v_3$ are linked to a vertex $v_2$. The indegree of the linked vertex should be more than 1 and here indegree($v_2$) = 2.

Each reasoning strategy s is a member of the set of reasoning strategies $S$. Functions are defined individually for each reasoning strategy which are accordingly called during the node append for graph creation.

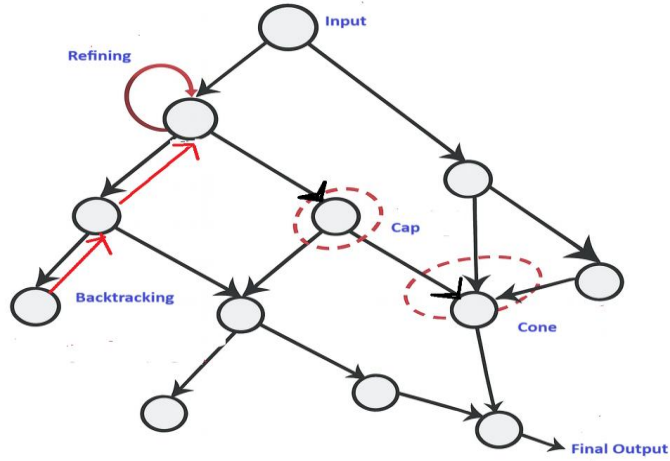

Figure 4. Directed Temporal Graph with examples of Reasoning Strategies

It is necessary to consider both knowledge and reasoning capabilities while answering a medical question or a query, as medical problems require richer domain knowledge and knowledge accuracy (Wu et al. 2025b). Moreover, the dynamic retrieval of external knowledge during inference can help to get the updated information of an evolutionary domain like healthcare. Hence, for a given input query q, first an external retrieval system R extracts knowledge K from a vast knowledge base such that K = R(q). Given q and R(q), MLLM synthesizes domain knowledge $k$ by integrating its intrinsic parametric knowledge with external inputs, following the conditional distribution $k \sim p( k|q, R(q))$. Within the domain-specific context, MLLM generates reasoning steps $r$ conditioned on q, R(q), and the integrated knowledge k, adhering to the reasoning distribution $r \sim p( r|q, R(q), k)$. During any iterative step of a chain of reasoning, an answer $a_i$ is based on the concatenation of knowledge $k_i$ and reasoning $r_i$ (Wang et al. 2025c). Each reasoning step $r_i$ has a corresponding integrated knowledge point $k_i$ during the iterative reasoning refinement process of temporal graph-based reasoning.

Finally, MLLM generates a node v based on a reasoning strategy and appends it to the sequence of nodes or a path of reasons. A verification is done for each generated node to assess whether the reasoning is correct or further refinement required by comparing with the ground-truth answer. A medical verifier similar to one proposed in (Chen et al., 2024b) can be utilized for this purpose. The verifier checks the reasoning answer $a$ with ground-truth answer $a^*$ to provide binary feedback, 'True' or 'False', such that Verifier (a, a$^*$) $\epsilon$ {True, False}. The iteration process continues till the correct answer is encountered and the corresponding node becomes the final node. The query $q$ with the final reason $r_f$ and answer $a_f$ are stored in the SFT dataset, $D_{SFT}$ (for Supervised Fine-Tunning in the training phase), while $q$, $r_f$, $a_f$ and the corresponding initial time $t_0$ and final time $t_f$ are stored in the temporal dataset $D_{TEMP}$ for future analysis and diagnosis.

The entire graph reasoning process is summarized in Algorithm 1.

---

**Algorithm 1 :** Graph-based Temporal Medical Reasoning

---

**Input** : Multimodal Input i $\epsilon$ $I$ (text, image, video, audio..), Query, q $\epsilon$ $Q$, Trained (SFT + RL) Medical Verifiable Problems, $M$ = {(q, a$_i^*$)}, MLLM (GPT-4o), Reasoning Strategies $S$, and $N$, maximum number of iterations for model refinement

V ← {Φ}, E ← {Φ}, T ← {Φ}, G ← (V, E, T)          // initializing temporal graph G

r$_f$ ← {Φ},  a$_f$ ← {Φ}, t$_0$ ← {Φ}, t$_f$ ← {Φ}          // initializing final reason r$_f$ , final answer a$_f$ , final time t$_f$
                                                                                    // and initial time t$_0$

$D_{TEMP}$ ← {Φ}                                          // initializing temporal reasoning dataset $D_{TEMP}$

$M_R$ ← Break($M$)                                      // 'Break' function breaks $M$ into dataset $M_R$ for Reasoning

**for** (q, a$^*$) $\epsilon$ $M_R$ **do**
   V ← {Φ}, E ← {Φ}, T ← {Φ}, G ← (V, E, T)          // initializing temporal graph G
   r$_f$ ←{Φ},  a$_f$ ← {Φ}, t$_f$←{Φ}                      // initializing final reason r$_f$ , final answer a$_f$ and final time t$_f$
   K ← R(q)                                          // External retrieval function R extracts knowledge R(q) for input q
   **for** j ← 1 **to** L **do**                          // Maximum number of pre-set retry limit L for each (q, a$^*$)
      **for** i ← 0 **to** N-1 **do**                  // N is the maximum pre-set iteration count

$k_i \sim p(\,k_i \mid q, R(q))$

$r_i \sim p(\,r_i \mid q, R(q), k_i)$

**if** i = 0 **then**

    $(\{r_i, a_i\}, \{\Phi\}, \{t_i\}) \leftarrow \text{MLLM}_{\text{init}}(q, G)$      // Initial node creation for the graph G, MLLM is a mapping

    // function

**else**

    $s_i \sim S$

    $(\{r_i, a_i\}, E(V, \{v_i\}), \{t_i\}) \leftarrow \text{MLLM}_{S_i}(q, G)$   // New node, edge and corresponding time generation

**if** Verifier $(a_i, a^*)$ **then**

    $r_f \leftarrow r_i$

    $a_f \leftarrow a_i$

    $D_{\text{TEMP}} \leftarrow D_{\text{TEMP}} \cup \{(i, q, r_f, a_f, t_0, t_f)\}$

    **break**

**Output :** $r_f, a_f, t_0, t_f, G$

### 5.3.1.2 Performance Evaluation

There can be several metrics for evaluating the performance of temporal graph-based medical reasoning MLLMs. However, four important metrics, *Latency, Accuracy, Efficiency* and *Volume,* are considered here. High accuracy, high efficiency, low latency and high volume are generally desired in multimodal medical reasoning. However, practical implementations often require a balanced approach to optimize performance based on the specific context.

**Latency -** Latency is the time it takes for an MLLM to process input data (text, images, audio etc.) and generate the output (response). It is a measure of how quickly the model can "think" and produce output. Hence, a lower latency is preferred in MLLMs. Rapid decision-making with accuracy is crucial in healthcare. Low latency ensures that AI systems can process information quickly and provide timely insights to healthcare professionals, potentially impacting patient outcomes in critical situations.

    In the case of temporal graph reasoning, the latency can be measured by the total reasoning time to reach the final answer from the initial input query. This can be evaluated by computing the time difference between the final node and the initial (root) node of the final reasoning path. If $\tau$ represents latency, then $\tau$ can be measured through the expression

$$\tau = t_f - t_0$$

**Accuracy –** Accuracy measures the ratio between the correct predicted answer that matches the ground-truth answer to the total number of candidate answers for a given medical query with one or more multimodal inputs.

If $\mathcal{A}$ be the accuracy, then $\mathcal{A}$ is calculated as

$$\mathcal{A} = \frac{1}{N} \sum_i^N F(y_i, y^*)$$

where N is the total number of test cases, $y_i$ is the is the model's predicted answer index, $y^*$ is the ground truth answer and F is the function that returns '1' if the predicted answer matches the ground truth i.e. $y_i = y^*$ and '0' otherwise.

**Efficiency –** The ratio between the accuracy of the reason and the duration of the reason can measure the Efficiency of a reason. Efficiency is high if accuracy is high and reasoning duration is low. Hence, efficiency increases as accuracy increases, even if reasoning duration is not low.

    Let $r_i$ be any arbitrary reason, $t_i$ be the time at which the reasoning begins and node corresponding to $r_i$ is created, $t_{i+1}$ be the time at which the reasoning ends and next node corresponding to $r_{i+1}$ is created and A is the reasoning accuracy.

Then the reasoning duration is given by $\Delta_i = t_{i+1} - t_i$ and efficiency, $\mathcal{E}_i$ is calculated as

$$\mathcal{E}_t = A_i / \Delta_i$$

**Volume -** Multimodal medical reasoning can correctly derive the final result through a deep and comprehensive reasoning process. This requires consideration of a larger number of reasons or thoughts before the final reason. Hence, the *volume* can be defined as the number of preceding reasons from which there exist paths to the node representing the final reason in the temporal graph of reasons. These preceding reasons have an impact or lead to the said reason. If 'V' represents volume and 'n', the number of reasons, then $V = n$.

Hence for larger values of $n$ or higher volume, the impact of the graph of reasons is much higher. In real-life scenarios like medical domain, a greater number of correct reasons usually leads to better final results. So, this justifies considering a higher number of valid reasons or a larger $n$.

## 6 EXPERIMENTS

Probably being the first of its kind, this preliminary work on temporal medical reasoning places more emphasis on the analysis of reasoning patterns at different time points during a period under consideration for change or without change in multimodal inputs. It requires experiments to be carried out to study the reasoning accuracy at different time points due to updates or changes in medical domain knowledge, or due to changes in patients' health conditions, as reflected through multimodal inputs. Moreover, the evolution of patients' health conditions, including reasoning and diagnosis, can be analyzed over different periods. It is not yet possible to carry out a thorough benchmarking evaluation primarily due to the unique nature of the current work and the non-availability of real-life temporal medical reasoning data. The benchmarking effort will be carried out and reported in the future, subject to the availability of the said temporal data.

### 6.1 Experimental Details

#### 6.1.1 Datasets

A synthetic multimodal medical reasoning dataset was constructed from publicly available datasets of MedXpertQA (Zuo et al., 2025), ReasonMed (Sun et al.,2025), MultiCaRe (MultiCaRe, 2025), MedTrinity-25M (Xie et al., 2025), MMIST-ccRCC (Mota et al., 2024), PubMedVision (Chen et al., 2024a), QuiltLlaVA Instruct (Seyfioglu et al., 2025) and MIMIC-CXR (Johnson et al., 2024; Johnson et al., 2019), covering common medical question-answering datasets that vary in question complexity. A synthetic benchmark-style table of 10 sample multimodal medical reasoning questions is presented in Table 1, along with other relevant attributes. It covers diverse medical data with different focus areas and modality types.

#### 6.1.2 Baseline Models

The current model can be evaluated and compared with closed-source models GPT-4o, Claude-3.5-Sonnet, Gemini 2.5 Flash and open-source models Qwen2.5-VL, InternVL3 and LLaMA Vision 3.2.

### 6.2 Main Results

The temporal reasoning performance of different multimodal datasets in the proposed framework is presented only in the Table 1, using the evaluation metrics outlined in Section 5.3.1.2. Detailed analysis had been carried out to justify the validity of the proposed framework.

| Dataset | Focus Area | Modality Type | Accuracy | Reasoning Time (sec) |
|---------|-----------|---------------|----------|----------------------|
| ReasonMed | Pneumonia detection | Image + Text | 0.93 | 20 |
| MultiCaRe | Diabetes diagnosis | Text + Lab | 0.91 | 18 |
| MedXpertQA | Drug interaction | Text (EHR + DB) | 0.94 | 16 |
| MedTrinity-25M | Stroke detection | Image + Text | 0.90 | 24 |
| MMIST-ccRCC | Kidney cancer subtype | Image | 0.92 | 29 |
| PubMedVision | Literature-grounded reasoning | Text + Image | 0.89 | 27 |
| QuiltLlaVA | Surgical step recognition | Video + Audio | 0.90 | 28 |
| MIMIC-CXR | Cardiomegaly detection | Image + Text | 0.95 | 15 |
| MultiCaRe | ICU mortality prediction | Text + Lab + Vitals | 0.88 | 22 |
| MedTrinity-25M | COVID-19 severity | Image + Text | 0.94 | 25 |

Table 1. The temporal reasoning performance of multimodal medical datasets.

The Key Insights from Table 1 are:

*Coverage of modalities:*

- **Text** (EHR, clinical notes, PubMed articles)
- **Image** (X-rays, CT, histopath slides, figures)
- **Video** (surgical and endoscopic recordings)
- **Audio** (narrations, instructions)

*Accuracy (ratio form)*
- Ranges from 0.88 to 0.95

*Reasoning Time*
- Varies from $15 - 29$ seconds, depending upon the reasoning complexity.

Few experiments were conducted using various evaluation metrics, multimodal types, and their individual or combined changes with time. The primary objectives were to observe the accuracies, durations, and efficiencies of reasoning with their variations over different periods, the impact of changes in input multimodal data for different time intervals and the impact of the number of agents. The results with corresponding analysis are presented in the next section graphically through line charts and bar graphs for better understanding. Moreover, the impact of the number of agents in an adaptive group setting was also studied and analyzed during different periods.

### 6.3 Analysis

The analysis results of the temporal reasoning performance evaluation of medical tasks mentioned in Table 1 are illustrated step-by-step for better understanding. Each type of analysis, with a brief explanation of the results, is provided here.

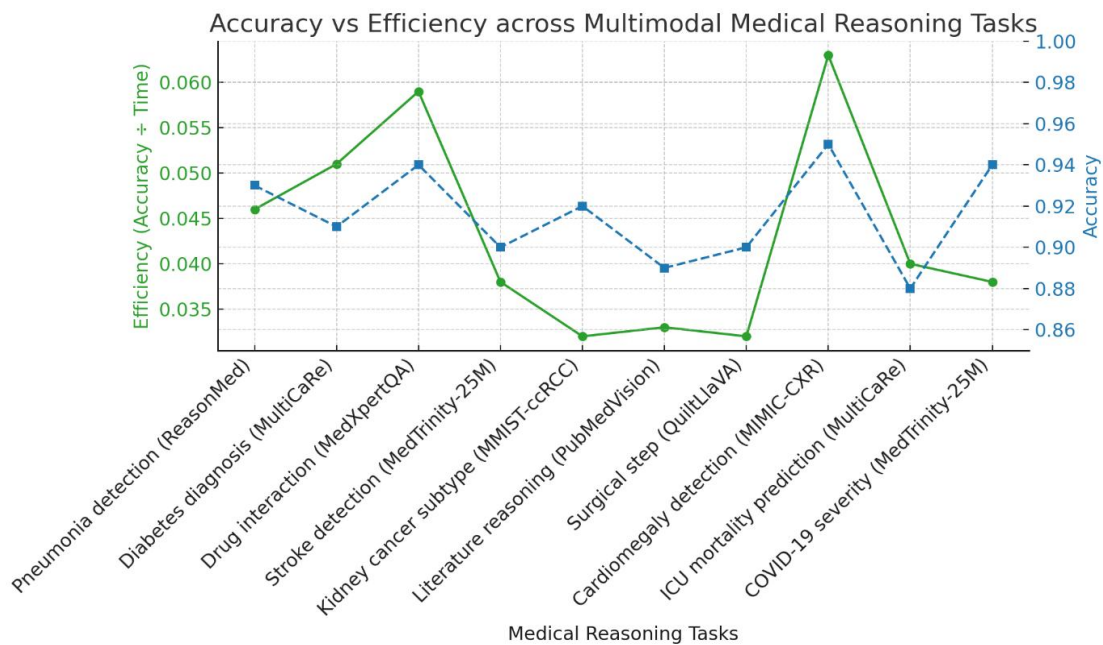

Figure 5. Accuracy vs Efficiency across Multimodal Medical Reasoning Tasks

In Figure 5, a dual-axis line chart (Accuracy vs Efficiency) is created to show how performance and efficiency trade off for these tasks. Here's the dual-axis line chart showing how Efficiency (green) and Accuracy (blue dashed) vary across the 10 multimodal medical reasoning tasks. It clearly illustrates that some tasks (e.g., Kidney cancer subtype, Surgical step recognition) have high accuracy but low efficiency due to long reasoning times, while others (e.g., Cardiomegaly detection, Drug interaction) achieve both high accuracy and efficiency.

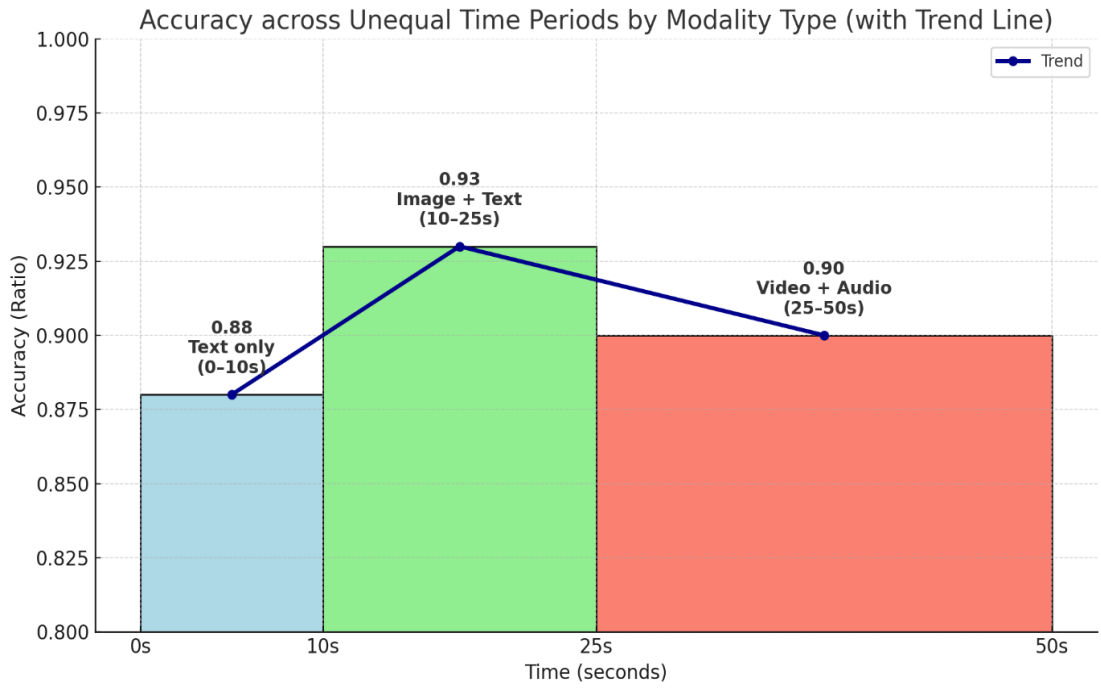

Figure 6. Comparative Accuracies across Unequal Time Periods by Modality Type

It has been usually observed that the type and number of modalities increase with the complex medical tasks due to the complexities in disease symptoms or conditions of the patient over different periods. This fact, along with the corresponding accuracies, has been illustrated in Figure 6. Here, a comparative bar chart was created for the modality types to show how accuracy changes during different unequal periods due to changes in modality types during those periods. A trend line (connecting period midpoints at 5s, 17.5s, and 37.5s) on top of the non-overlapping time-span bars added to show how accuracy evolves.

Next, the previous data was extended with modalities text + image + video, and an additional period of 25 seconds to see how accuracy and efficiency evolve further during this period. The new comparative bar chart in Figure 7 below illustrates the change.

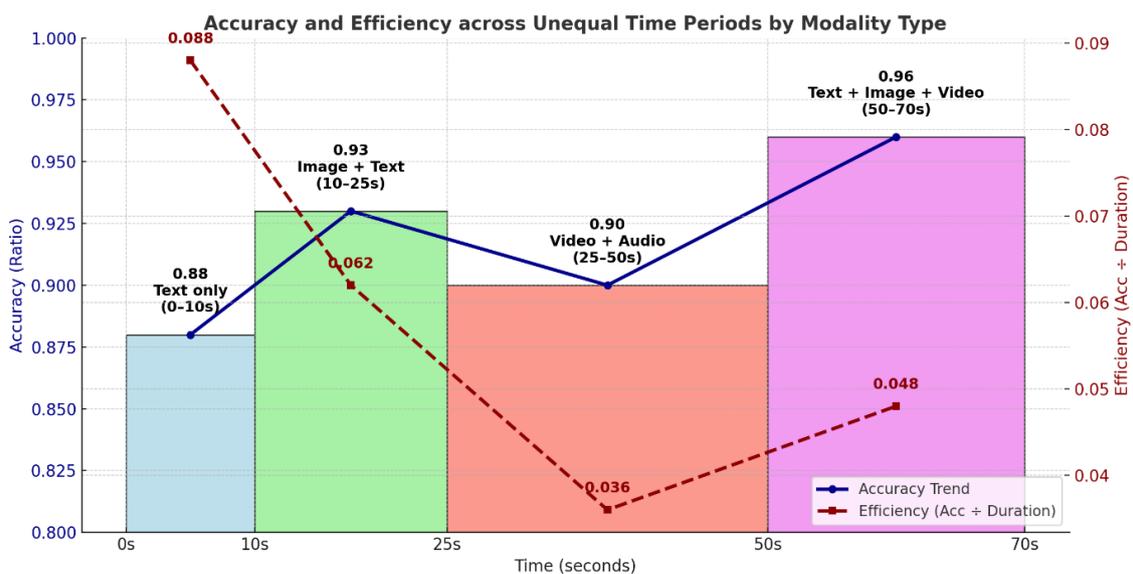

Figure 7. Comparative Accuracies and Efficiencies across Unequal Time Periods by Modality Type

It can now be clearly seen how accuracy peaks with Text+Image+Video, while efficiency fluctuates due to the longer reasoning duration.

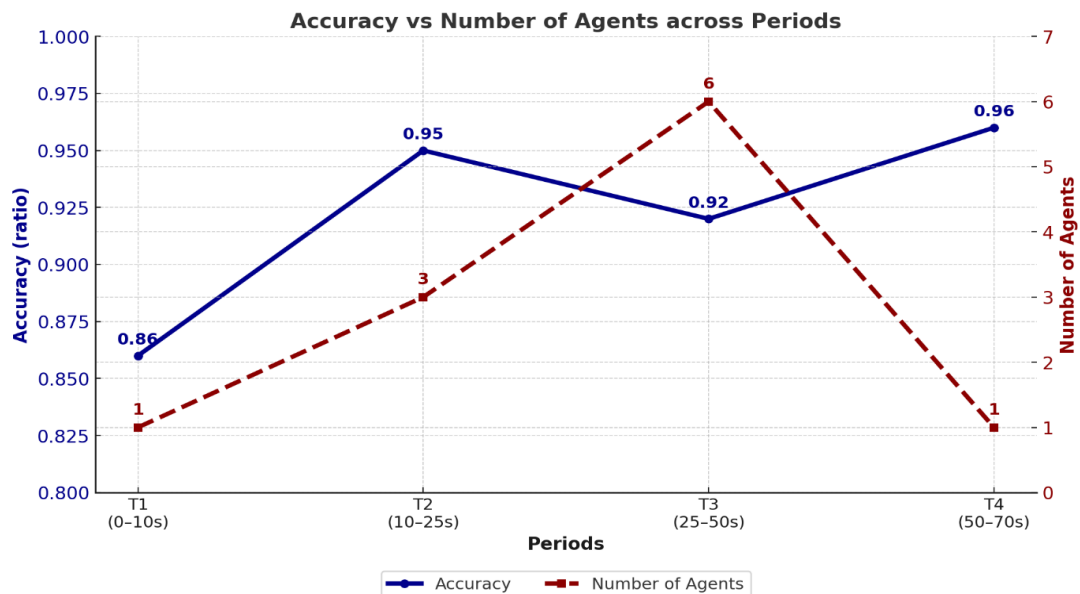

Figure 8. Comparative Accuracies for Variable Number of Agents across Periods

The number of agents involved in medical diagnosis varies according to the severity of the cases. Usually, the number of agents starts from one for GMP, where the first agent (GMP) checks the complexity of the case and can accordingly refer to other relevant expert agents for analysis during different periods for which the number of agents increase or decrease but finally drop down again to one for the final decision-making agent, e.g Primary Doctor agent who synthesizes reports from all agents and may include own opinion to provide the final accurate answer. Again, it has been observed that continuously involving a higher number of collaborating agents cannot always lead to better performance. Rather, a lower value may result in optimal performance (Kim et al., 2024). The line chart of Figure 8 clearly illustrates this fact.

## 7  CONCLUSIONS

The current paper has attempted a novel approach by incorporating a *temporal graph of reasons* through a temporal graph-based multi-agent reasoning framework to enhance the reasoning capabilities of multimodal language models in the healthcare and medical domain for better diagnosis and medical decision-making. The graphs with the reasonings and respective initial and final times created and stored at different time points during an investigation /analysis period for a patient can be utilized to assess the patient's conditions, diagnosis, and treatment planning for that period. Moreover, the reasoning graphs, final reasonings with initial and final times stored during different periods, can help to refer to the earlier archived reasonings for better understanding the current disease symptoms and diagnose accordingly. Finally, the reasoning times required by the medical queries posed to the agents can be utilized to assess different metrics for analysis. The proposed method first checks the severity of a patient case and accordingly assigns different domain-specific doctor agents. Each of these agents analyze various types of medical data and generates corresponding temporal reasoning paths step-by-step. The factual correctness of each reasoning step is additionally done by verifying extracted knowledge against extracted ground-truth sources. Then, the verification is done on the accuracy of the reasoning data before generating the final answer to ensure quality. Further cross-verification mechanism is introduced through the sharing of reasoning by multiple agents to update the reasoning process, leading to a final multi-agent collaborative diagnosis result after approval by the authorized doctor agent. This explicit multi-agent temporal reasoning framework with multiple levels of verification helps to enhance the reasoning potential in complex healthcare scenarios and leads towards accurate answers.

## 8  LIMITATIONS AND FUTURE WORKS

While providing promising directions of the proposed framework for evaluating temporal reasoning capabilities in healthcare domain, several important limitations remain. These are:

- Constructed a synthetic multimodal medical reasoning dataset from several well-curated multimodal datasets, mainly covering multimodal medical question-answers that vary in question complexity. Future work will focus on building a multimodal dataset with more extensive medical knowledge and complex medical reasoning capabilities.
- Due to the non-availability of real-life temporal medical reasoning data and earlier multimodal temporal reasoning models, the benchmarking with other works could not be carried out and will be covered in future work, subject to the availability of such real-life data.
- Though there are several stages of verification in the proposed framework, there is a lack of systematic verification of model outputs by practicing clinicians for final confirmation. This raises concerns about the reliability and alignment of model-generated reasoning paths and diagnosis by the multi-agents with established medical knowledge. Future work needs to establish a more rigorous verification process involving manual domain experts to assess answer correctness, the validity of reasoning steps, and potential hallucinations.
- A deeper understanding of the effective application of RLVR in medical contexts is further needed.
- Although the current preliminary work on temporal medical reasoning focused on multi-agent and ensemble approaches in temporal multimodal reasoning, a deeper exploration of the ensemble methods, like step-wise verification, task-wise verification, and dynamic agent collaboration, needs to be done for more effective medical reasoning systems.

This work marks a preliminary effort on temporal reasoning in the healthcare domain. Future developments will attempt to focus on real-life, improved, and diversified medical datasets and benchmarking with temporal reasoning models subject to the availability of such models.

# APPENDIX

## Few Case Study Examples

Here are two real-life medical examples from *Lingshu* (LASA Team et al., 2025), one from Elicit and Enhance *MedE2* (Mu et al., 2025) to illustrate medical reasoning and report generation and one from *MDAgents* (Kim et al., 2024) to illustrate a case of collaborative multi-agents for medical reasoning and decision making. Figure 9 presents the medical VQA cases of Lingshu across a distinct medical modality, X-Ray while Figure 10 presents the case study of Lingshu in generating a medical report. Figure 11 presents the reasoning example for reasoning and treatment by MedE2. Finally, an illustrative example of *MDAgents* in a moderate complexity case is presented in Figure 12.

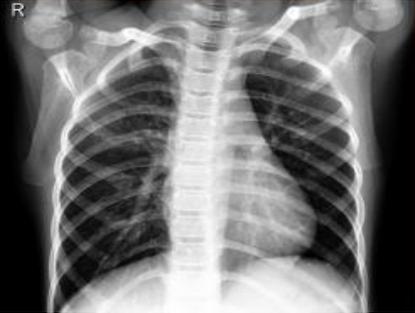

Figure 9.  Case study of Lingshu in visual question answering on X-Ray

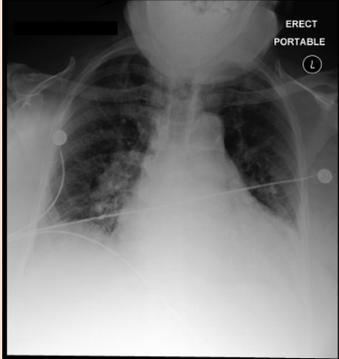

Figure 10.  Case study of *Lingshu* in generating Medical Report

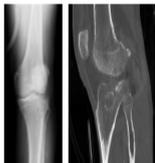

Figure 11. Reasoning for Diagnosis and Treatment using *MedE²*

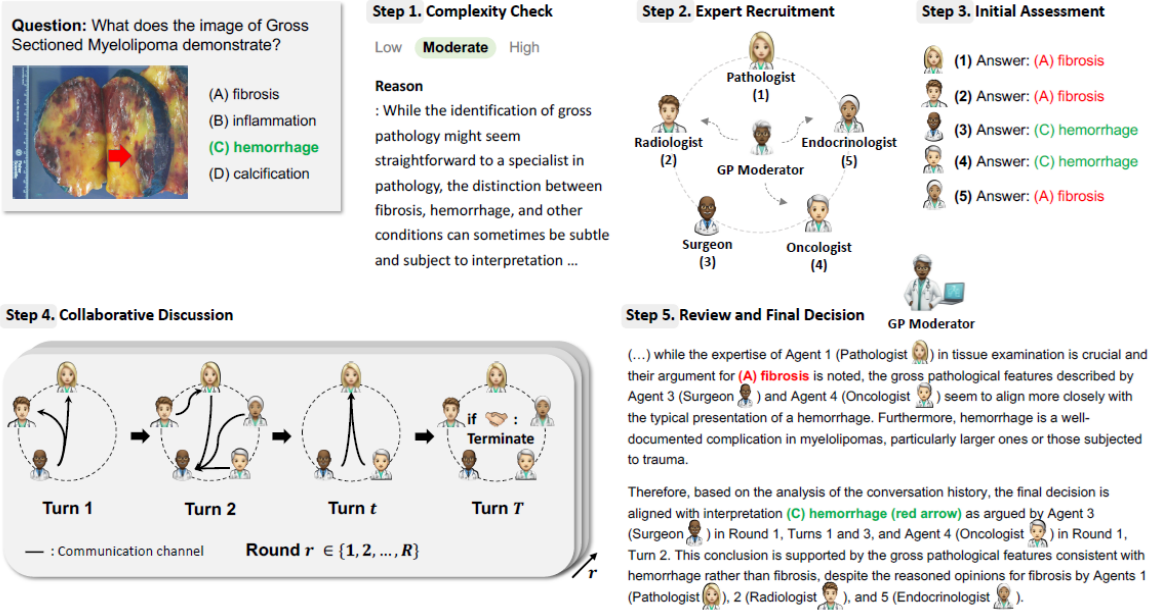

Figure 12. Example of MDAgents in a moderate complexity case from the PMC-VQA dataset.